\documentclass[notitlepage]{article}

\usepackage{natbib}
\usepackage{titlesec}
\usepackage[titletoc,toc,title]{appendix}
\titleformat{\subsection}{\normalfont\itshape}{\thesubsection.}{0.5em}{}
\titleformat{\subsubsection}{\normalfont\itshape}{\thesubsubsection.}{0.6em}{}
\usepackage{adjustbox}
\usepackage{booktabs,siunitx,rotating}
\usepackage{lineno}
\usepackage{pgf,tikz}
\usepackage{appendix}
\usepackage{mathrsfs}
\usepackage{amsmath}
\usepackage{multirow}
\usepackage{booktabs} 
\newcommand{\ra}[1]{\renewcommand{\arraystretch}{#1}}
\usepackage{mathtools}
\usepackage{array}
\usetikzlibrary{arrows}
\usepackage{pdflscape}
\usepackage{textcmds}
\usepackage{amssymb}
\usepackage{hyperref}
\usepackage{textcmds}
\usepackage{caption}
\usepackage[nottoc,notlot,notlof]{tocbibind}

\usepackage{etoolbox}
\usepackage{lipsum} 
\usepackage[utf8]{inputenc} % allow utf-8 input
\usepackage[T1]{fontenc}    % use 8-bit T1 fonts
\usepackage{hyperref}       % hyperlinks
\usepackage{url}            % simple URL typesetting
\usepackage{booktabs}       % professional-quality tables
\usepackage{amsfonts}       % blackboard math symbols
\usepackage{nicefrac}       % compact symbols for 1/2, etc.
\usepackage{microtype}      % microtypography
\usepackage{sidecap}
\usepackage{lscape}
\usepackage{wrapfig}
\usepackage{multirow}
\usepackage{enumitem}
\usepackage{graphicx}
\usepackage{dcolumn}
\usepackage{subcaption}
\usepackage{indentfirst}
\usepackage{amsmath, amssymb, mathtools, bm, amsthm}
\usepackage{amsfonts}
\usepackage{gensymb}
\usepackage{placeins}
\usepackage{booktabs} % for professional tables
\usepackage{todonotes}
\usepackage{hyperref}
\usepackage[ruled,vlined]{algorithm2e}
\usepackage{algorithmic}
\usepackage{bm}
\usepackage{cleveref}

\usepackage{pifont}% http://ctan.org/pkg/pifont
\title{Real Estate Property Valuation using Self-Supervised Vision Transformers}

\author{Mahdieh Yazdani and Maziar Raissi\\~\\ University of Colorado Boulder}

\date{}

\begin{document}

\maketitle

\begin{abstract}
The use of Artificial Intelligence (AI) in the real estate market has been growing in recent years. In this paper, we propose a new method for property valuation that utilizes self-supervised vision transformers, a recent breakthrough in computer vision and deep learning. Our proposed algorithm uses a combination of machine learning, computer vision and hedonic pricing models trained on real estate data to estimate the value of a given property. We collected and pre-processed a data set of real estate properties in the city of Boulder, Colorado and used it to train, validate and test our algorithm. Our data set consisted of qualitative images (including house interiors, exteriors, and street views) as well as quantitative features such as the number of bedrooms, bathrooms, square footage, lot square footage, property age, crime rates, and proximity to amenities. We evaluated the performance of our model using metrics such as Root Mean Squared Error (RMSE). Our findings indicate that these techniques are able to accurately predict the value of properties, with a low RMSE. The proposed algorithm outperforms traditional appraisal methods that do not leverage property images and has the potential to be used in real-world applications.\\

\noindent \textit{keywords}: housing price prediction model, hedonic model, self-supervised vision transformers, computer vision, deep neural networks, real estate property appraisal, regression analysis, image data.

\end{abstract}

\section{Introduction}
In many families, residential property is one of the most important components of a household’s wealth (see e.g., \cite{arvanitidis2014economics}). As a result, home prices are of great interest to both current and potential homeowners. The property prices are not only important to stakeholders, but also to insurance companies, property developers, appraisers, tax assessors, brokers, banks, mortgage lenders, and policy makers (see e.g., \cite{frew2003estimating} and \cite{yazdani2020individuals}). Therefore, accurate predictions and trend analyses in real estate prices can aid these groups in making informed decisions (see e.g., \cite{fan2018house}). For several decades, the estimation of real estate assets has relied on hedonic pricing models (see e.g., \cite{rosen1974hedonic}, \cite{del2017hedonic}, \cite{yazdani2021house}, and \cite{krol2020application}). Hedonic models are one of the most widely accepted methods for estimating house prices and are commonly used to recover the implicit prices of house attributes. The urban property markets contain high variation in the structural, locational, neighborhood, and environmental attributes. Most hedonic models employ quantitative features such as the number of bedrooms, bathrooms, square footage, lot square footage, property age, etc. to appraise house prices (see e.g., \cite{geng2015study}, \cite{lasota2011empirical}, and \cite{del2017hedonic}).\\

As alternative tools, various machine learning and deep learning algorithms such as artificial neural networks (ANN), $k$ nearest neighbor (kNN), bounded fuzzy possibilistic method (BFPM), random forest (RF), and support vector regression (SVR) have been proposed for house price valuation and real estate property price prediction models (see e.g., references in \cite{bigus1996data}, \cite{lenk1997high}, \cite{kauko2002capturing}, \cite{curry2002neural}, \cite{pagourtzi2003real}, \cite{limsombunchai2004house}, \cite{peterson2009neural}, \cite{zahra}, \cite{zhou2015predicting}, \cite{islam2009housing}, \cite{selim2011determinants}, \cite{morano2013bare}, \cite{yazdani2018comparative}, \cite{ali2015housing}, \cite{park2015using}, \cite{vceh2018estimating}, \cite{poursaeed2018vision}, \cite{hong2020house}, and \cite{pai2020using}). These studies have yielded mixed results. For example, \cite{selim2011determinants} compared the prediction performances of the hedonic price regression and ANN models for the prediction of dwelling prices in Turkey and found that ANN performed better. Similarly, \cite{yao2018mapping} integrated a convolutional neural network with RF to analyze the housing market in Shenzhen and found promising results for the application of machine learning and deep learning algorithms. \cite{park2015using} has compared the performance among several machine learning algorithms such as Repeated Incremental Pruning to Produce Error Reduction, Na{\"i}ve Bayesian, and AdaBoost to identify better forecasting models. This study has shown the promising application of machine learning and deep learning algorithms in housing markets. \cite{yazdani2021machine} compared the performance of artificial neural networks, random forest, and $k$ nearest neighbor approaches to that of hedonic method on house price prediction in the city of Boulder, Colorado. This study demonstrated that random forest and artificial neural networks algorithms can be better alternatives over the hedonic regression analysis for prediction of the house prices in the city of Boulder, Colorado. However, \cite{kontrimas2011mass} empirically studied several different models on structured features such as house type, size, house age, etc. Their findings indicate that linear regression surpasses that of neural network methods and linear regression may be a better alternative. \\

However, these models do not fully account for the complexity of the housing market decision-making process. Homebuyers take into account not only structural factors, socio-economic status of the neighborhood, environmental amenities, and location, but also evaluate the interior and exterior of properties such as appliances, house structure, etc. The visual appearance of houses, which is likely one of the most important factors in a buyer's decision, is often ignored in hedonic models. This could be partly due to lack of availability of house images or difficulty in quantifying visual content and incorporating it into hedonic methods. The real estate appraisal process could benefit from the introduction of images into the models as it can represent the overall house construction style and quality. The recent development of robust computer vision algorithms makes it possible to analyze unstructured data such as images. As images are high-dimensional, deep learning methods are needed to transform them into structured data. With deep learning, image features can be quantitatively described and included in appraisal models. \\

In image-related tasks, Convolutional Neural Networks (CNNs) such as LeNet [\cite{lecun1998gradient}], AlexNet [\cite{krizhevsky2017imagenet}], VGG [\cite{simonyan2014very}], Inception [\cite{szegedy2017inception}], ResNet [\cite{he2016deep}], DenseNet [\cite{huang2017densely}], Xception [\cite{chollet2017xception}], MobileNet [\cite{howard2017mobilenets}], and EfficientNet [\cite{tan2019efficientnet}] have been historically employed. However, with the advent of Vision Transformers ViT [\cite{dosovitskiy2020image}] there is a shift towards using transformer-based models for image related tasks. These models have shown to achieve state-of-the-art results on various benchmarks and are increasingly being adopted in industry as well. Vision Transformers are a recent breakthrough in computer vision (CV) and deep learning that have been shown to be superior to CNNs in some cases. Vision Transformers are able to learn more global features from images because of their self-attention mechanism and are therefore more effective for tasks such as transfer learning. Transfer learning is a machine learning technique that allows a model trained on one task to be applied to a different but related task. This is achieved by transferring the knowledge gained from one data set to another. This can be done by using the weights and biases of a pre-trained model as the starting point for training a new model, or by fine-tuning the pre-trained model on the new task. Transfer learning can save a significant amount of time and resources, as well as improve the performance of the new model. It is particularly useful when there is limited data available for the new task, as the pre-trained model can provide a good starting point for learning the new task (see \cite{raissi2023open}).\\

In this paper, we propose a novel approach to property valuation that leverages the power of self-supervised vision transformers, a recent breakthrough in computer vision and deep learning. Self-supervised deep learning enables the model to learn features from raw data without the need for manual annotations. This means that the model is able to learn from a wider range of data and can discover more general and abstract features. In contrast, supervised learning relies on labeled data which is typically more limited and specific to the task it was labeled for. Additionally, self-supervised learning methods can learn the underlying structure of the data, which can be useful for a wide range of tasks. The representations learned from self-supervised learning are learned from the data itself and are not dependent on the specific task, allowing them to be more generalizable. This can make the features learned through self-supervised deep learning transfer better to new tasks than features learned through supervised learning.\\

Our algorithm leverages self-supervised vision transformers from the computer vision literature to perform transfer learning and extract quantitative features from qualitative images. This enables us to combine machine learning, computer vision, and hedonic pricing models, all trained on a data set of real estate properties from Boulder, Colorado. This data set includes both qualitative images and quantitative features such as structural factors, socio-economic status of the neighborhood, environmental amenities, and location. We evaluate the performance of our model using metrics such as Root Mean Squared Error (RMSE), and our results show that this technique can accurately predict property values with a low RMSE. In summary, this paper presents a new method for property valuation that utilizes self-supervised vision transformers and outperforms traditional appraisal methods that do not incorporate property images, making it a valuable tool for real-world applications.\\

The paper is organized as follows. Section 2 showcases the data set, collected by the authors, which is novel and unique. In Section 3, we provide an overview of the machine learning, computer vision, and hedonic pricing models applied. Section 4 discusses the results obtained from these techniques. Finally, Section 5 offers conclusions and implications.

\section{Data}

This study incorporates both qualitative images and quantitative features to enhance the accuracy of the house price prediction models. The real estate data sets used were collected from various sources, including Multiple Listing Service databases\footnote{\url{https://realtyna.com/blog/list-mls-us}}, Public School Ratings\footnote{\url{https://www.greatschools.org}}, Colorado Crime Rates and Statistics Information\footnote{\url{https://www.neighborhoodscout.com/co/crime}}, CrimeReports\footnote{\url{https://www.cityprotect.com}}, WalkScore\footnote{\url{https://www.walkscore.com}}, Street View \footnote{\url{https://www.instantstreetview.com}}, recolorado\footnote{\url{https://www.recolorado.com}} and US Census Bureau\footnote{\url{https://data.census.gov/cedsci}}. We merged all data sets obtained from various websites. To isolate the influence of time on property prices, the data used in this study is restricted to houses sold in a single year between January 1, 2019 and December 31, 2019 (see \cite{eckert1990property}). Our collected data set consists of $1061$ residential properties sold in the city of Boulder, Colorado in 2019. During the screening process, we determined that four of the properties were in poor condition and in need of rebuilding, so we removed those four observations. Additionally, we excluded the only furnished property, which was sold with a lot of luxury furnishings, among all the transactions, which were sold unfurnished. Records associated with 30 reported horse properties and 4 duplicate transactions were also eliminated. \\

In our screening process, we encountered missing data points for various variables such as the number of bedrooms, bathrooms, parking, Lot Area, HOA fees, Solar Power, and Pool, Bathtub, Sauna, or Jacuzzi. To address these missing values, we updated some of the data by visiting different websites. However, a few observations still had missing data points for the continuous variable Lot Area and the dummy variables Solar Power and Pool, Bathtub, Sauna, or Jacuzzi. To avoid sample size reduction and sample selection bias (see e.g., \cite{hill2013hedonic}), we chose to impute the missing values with the mean for the Lot Area and the mode for the dummy variables. We also identified outliers and applied Winsorization to reduce their impact on the analysis. The data cleaning process left us with a sample of $1018$ observations. Descriptive statistics for the variables in this data set, are summarized in Tables \ref{table1} and \ref{table2}.  These statistics include the mean, standard deviation, minimum and maximum values, as well as the relative standard deviation (the coefficient of variation), which represents the extent of variability in relation to the average of the variable.\\

\begin{table}[h!]
\caption{Descriptive Statistics for Numerical Variables.}\label{Table1}
\centering
\begin{adjustbox}{max width=\textwidth}
\begin{tabular}{@{}l l l l l l@{}}\hline\hline\toprule\\
\multirow{2}[3]{*}{Variables} & \multicolumn{5}{c}{Aggregate Level}  \\\\
\cmidrule(lr){2-6}\\

&Mean & St. Dev.&Min &Max & Coefficient of Variation \\ [0.5ex] % 
\hline
\hline
\\
$\text{House Price}$ ($\$$)& $896,332$ & 679,195.8 & 112,897 & 7,200,000 &$76\%$\\

% $\text{Ln} (\text{House Price})$& 13.49 & 0.57 & 12.18 & 14.57&$4\%$\\

% $\text{Ln} (\text{Lot Area})$ (SqFt) & 6.21 & 4.37 & 0.00 & 14.27 & 70\%\\

% $(\text{Ln} (\text{Lot Area}))^2$ & 57.67 & 43.97 & 0.00 & 203.68& 76\% \\ 

$\text{Lot Area}$ (SqFt)& 18,367 & 84,909.71 & 0 & 1,577,744   &462.29\%\\

$\text{Living Area}$ (SqFt)& 2,264 & 1,398.79 & 416 & 10,354   &61.78\%\\

% $(\text{Ln}(\text{Living Area})$ (SqFt)& 7.54 & 0.61 & 6.03 & 9.25  &8\%\\

% $(\text{Ln} (\text{Living Area}))^2$ & 57.27 & 9.24 & 36.37 & 85.47&16\%\\

Age (year) & 43.12 & 21.46 & 1 & 98&50\%\\

% $\text{Age}^2$  & 2,319.19 & 2,253.79 & 1 & 9,604 & 97\%\\

%Number of Bedrooms  & 3.16 & 1.17 & 0 & 7& Numerical \\

Full Bathroom  & 1.55 & 0.76 & 0 & 3& 49\% \\

Half Bathroom  & 0.41 & 0.51 & 0 & 2& 124\%\\ 

$\frac{3}{4}$ Bathroom  & 0.64 & 0.69 & 0 & 2& 108\%\\

Parking & 1.68 & 0.71 & 0 & 3& 42\%  \\ 

HOA Fees (annually) ($\$$)  & 1,693.32 & 2,033.16 & 0 & 7,113& 120\%\\ 

Drive to CBD (minute)  & 11.42 & 6.91 & 1 & 26& 61\% \\

Walk to E.School (minute) & 21.37 & 17.31 & 2 & 68& 81\% \\ 

Walk to M.school (minute) & 33.21 & 27.72 & 2 & 96  & 83\%\\ 

Walk to H.school (minute) & 46.94 & 32.92 & 4 & 122 & 70\%\\ 

Married ($\%$) & 42.97 & 16.87 & 9.90 & 70.30& 40\%\\

Median Household Inc. ($\$$)  & 61,137.44 & 20,891.56 & 19,985 & 96,406&  34\%\\

Neighborhood's Population & 43,641.85 & 46,872.42 & 888 & 99,081 & 107\%\\\\
\hline
Sample size  & 1018   \\ 
 [1ex]

\hline \hline
\end{tabular} %You can adjust how far below the table the text should appear
\end{adjustbox}
\label{table1}
\end{table}

\begin{table}[h!]
\caption{Descriptive Statistics for Categorical Variables.}\label{Table2}
\centering
\begin{adjustbox}{max width=\textwidth}
\begin{tabular}{@{}l l l l  l@{}}\hline\hline \toprule\\
\multirow{2}[3]{*}{Variables} & \multicolumn{4}{c}{Citywide Level}  \\\\
\cmidrule(lr){2-5}\\

& Levels& Description & Frequency & Percent \\ [0.5ex] % inserts table %heading
\hline
\hline\\
& 0 &No& 658 &64.64 \\[-1ex]  

\raisebox{1ex}{Pool, Bathtub, Sauna, or Jacuzzi} &  1&Yes&360&35.36 \\[1.5ex]

& 0&No & 724 &71.12\\[-1ex]  

\raisebox{1ex}{Solar Power} &  1&Yes &294&28.88\\[1.5ex]
 
& 1&A & 334 &32.81  \\[-1ex]  

\raisebox{1ex}{Nearest E.School Rank} & 2  &B
&548 &53.83  \\

& 3 &C &136 &13.36\\[1.5ex]
 
& 1 &A& 125&12.28 \\[-1ex]  

\raisebox{1ex}{Nearest M.School Rank } & 2  &B
&633 &62.18 \\

& 3& C &260 &25.54 \\[1.5ex]
 
\raisebox{1ex}{Nearest H.School Rank} & 1 &A
&1018 &100 \\[1.5ex]

& 1&Central & 230&22.59 \\[.01ex] 

& 2&North 
&238 &23.38  \\[-.05ex] 
& 3& South &143 &14.05 \\[-1ex] 
\raisebox{1ex}{Region} & 4& East&200 &19.65 \\

& 5& Gunbarrel&125 &12.28 \\

& 6& Rural&82 &8.06 \\[1.5ex]

& 0 &0 bedroom & 3&0.29\\[.01ex] 

& 1 &1 bedroom & 72&7.07\\[-.05ex] 

& 2&2 bedrooms &253 &24.85 \\ [-1ex] 

\raisebox{1ex}{Number of Bedrooms} 
& 3& 3 bedrooms&285 &28.00 \\

& 4& 4 bedrooms&249 &24.46 \\

& 5& 5 bedrooms&127 &12.48 \\

& 6& 6 bedrooms&27 &2.65 \\

& 7& 7 bedrooms&2 &0.20 \\[2.4ex]

& 1 &Condominum & 324&31.83 \\[-1ex]  

\raisebox{1ex}{Property Type} & 2&Town - Home 
&90 &8.84  \\

& 3& Single-Family &604 &59.33 \\[1.5ex]

& 1 &Highest crime rate&82&8.06\\[-.75ex]  

\raisebox{1ex}{Neighborhood's Crime Level} & 2  &Middle crime rate&505&49.61\\

& 3& Lowest crime rate & 431 &42.34\\[1.75ex]
 
 \hline
Sample size&-&1018&100&-\\

\hline \hline
\end{tabular} %[2.3] %You can adjust how far below the table the text should appear
\end{adjustbox}
\label{table2}
\end{table}

\newpage

\begin{figure}[h!]
\centering
\includegraphics[width=\textwidth]{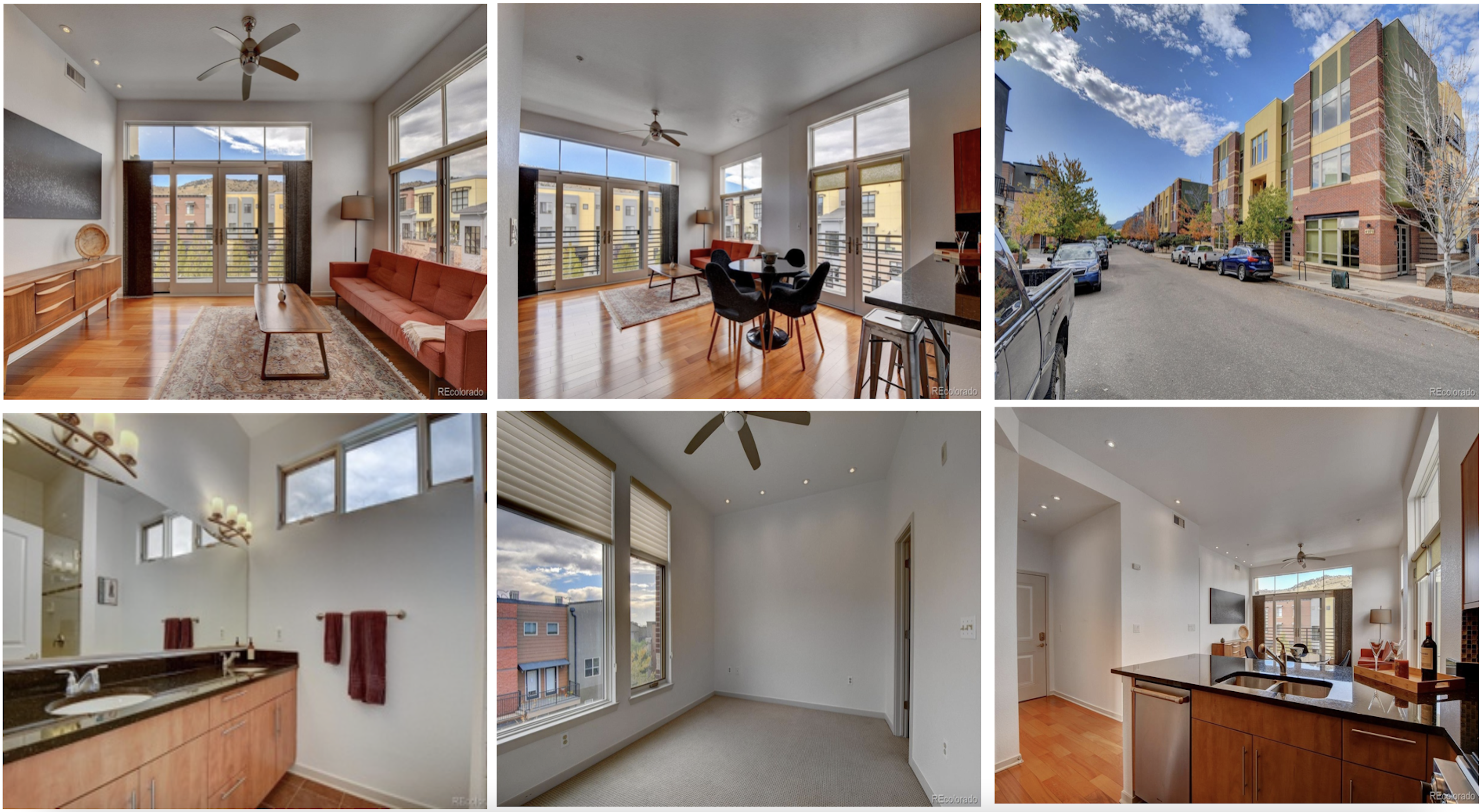}
\caption{Some sample images from a representative property listing in our data set.}\label{sample_images}
\end{figure}

Our data set also included property images. The images include detailed interior shots of rooms like the living room, dining room, bedrooms, and bathrooms, as well as exterior images showcasing the architectural style, texture of the building materials, and the design of windows and doors. Additionally, we also have street view images which give us a sense of the surrounding neighborhood and the overall aesthetic of the area. Some sample images from a representative property listing in our data set are depicted in Fig.~\ref{sample_images}. It is worth mentioning that the number of images per listing can vary and 14 properties were found to have no accompanying images. These were excluded from the data set, reducing the final sample size from 1018 to 1004.

\begin{figure}[h!]
\centering
\includegraphics[width=\textwidth]{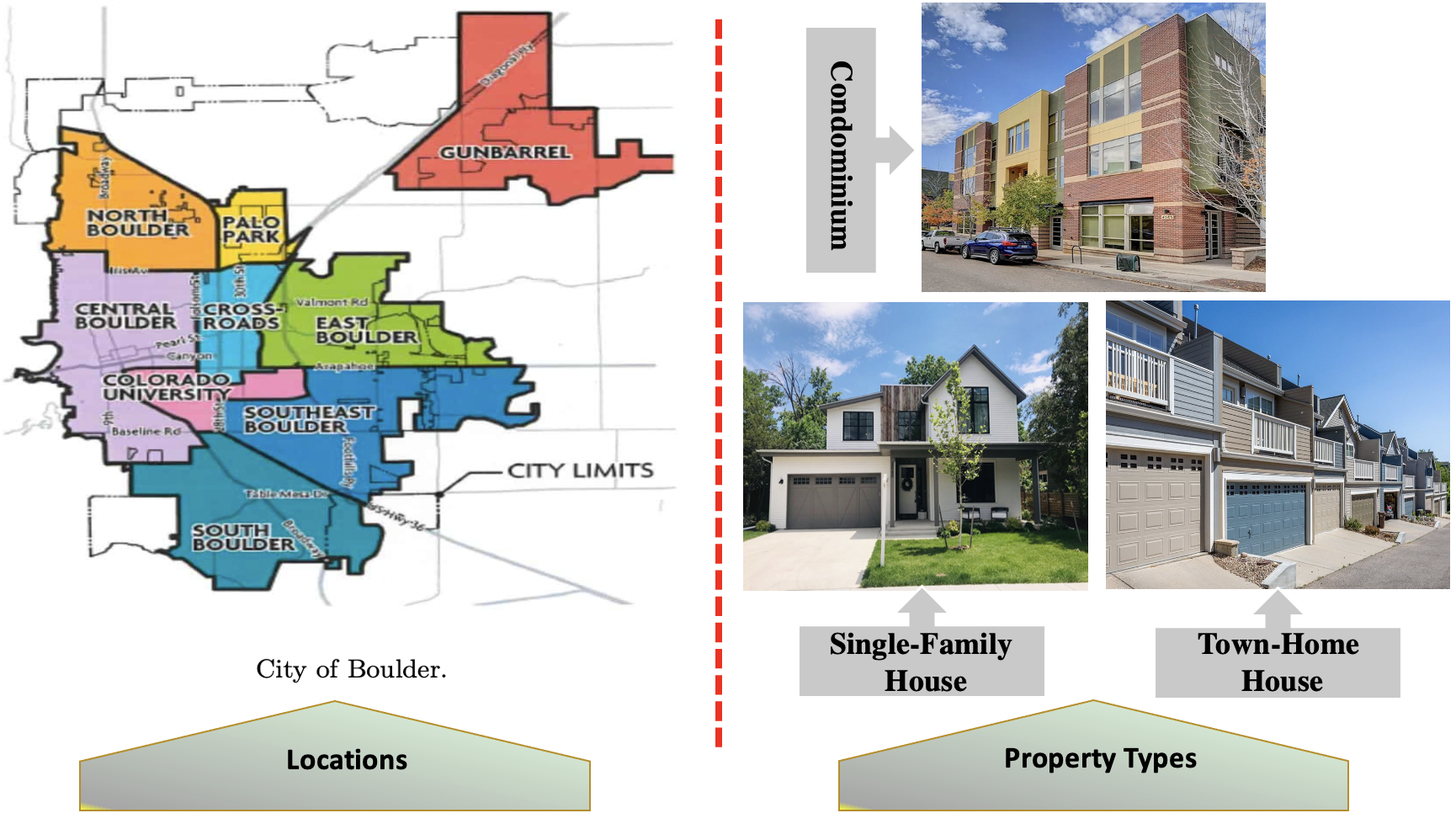}
\caption{City of Boulder on the map and the various property types.}\label{location_type}
\end{figure}

City of Boulder is divided into seven different geographical locations; central Boulder, downtown Boulder, old north Boulder, north Boulder, south Boulder, east Boulder, Gunbarel, and rural areas. With city development, the old north Boulder neighborhood is in central Boulder. We explored the geographical location of each property by making use of Google Maps. The property types in the housing market in the city of Boulder are classified as condominiums, town-homes, and single-family houses. Figure \ref{location_type} plots the city of Boulder on the map and the property types.\\

The sample includes $604$ single-family houses, $324$ condominiums, and $90$ townhomes. Single-family properties range in price from $\$216,575$ to $\$7,200,000$ with an average price of $\$1,160,321$. Townhomes range from $\$115,000$ to $\$1,421,000$ with an average price of $\$627,960$, while condominiums range from $\$112,897$ to $\$2,600,000$ with an average price of $\$478,751$. On average, single-family homes have $3.82$ bedrooms, with $0.17\%$ having no bedroom and $24.51\%$ having $5$ or more. Townhomes have $2.97$ bedrooms on average, with $7.78\%$ having $5$ or more. Condominiums have $1.98$ bedrooms on average, with $0.62\%$ having no bedroom and $0.31\%$ having $5$ or more. $3.31\%$ of single-family homes have solar power and $18.05\%$ have a pool, bathtub, sauna, or jacuzzi. Information about solar power and amenities in townhomes and condominiums is limited.\\

In the locational submarkets of Boulder, the average age of dwellings varies from $30$ years in North Boulder to $59$ years in Central Boulder. In the spatial submarkets, the North Boulder region has 238 transactions with a house price range of $\$134,306$ to $\$4,500,000$ and an average price of $\$807,214$. The Central Boulder region has 230 transactions with a house price range of $\$115,000$ to $\$7,200,000$ and an average price of $\$1,256,235$. Table \ref{table:nonlin} provides more information about the descriptive statistics of the house prices in different submarkets. From Table \ref{table:nonlin} we learn that the deviation in residential property prices is lower in North, South, East, and Gunbarrel submarkets compared to the overall market level. However, the house price difference is higher in Central Boulder and rural areas. \\

\begin{table}[h!]
\caption{Summary Statistics of House Prices in Different Submarkets.\\}\label{Table5}
\captionsetup{singlelinecheck = false}
\centering %used for centering table
\ra{1.3}
\begin{adjustbox}{max width=\textwidth}
\begin{tabular}{@{}llllllllllllll@{}}\toprule
\hline
\multirow{2}[3]{*}{Market Level} & \multicolumn{13}{c}{House Price} \\ \cmidrule{6-14} 
&&&&&& Mean && St. Dev.&&   Min && Max \\ \midrule
\hline
Citywide  &&&&&& $896,332$  && $679,195.8$  && $112,897$ && $7,200,000$\\
Single-Family &&&&&&$1,160,321 $&&$739,401.9$ &&$216,575$&&$7,200,000$\\
Town-Home   &&&&&&$627,960$ &&$246,470$ &&$115,000$  && $1,421,000$\\
Condominum &&&&&& $478,751$ &&$299,644.7$&&$112,897$ &&$2,600,000 $\\[-3ex]
{\color{white}your comment ...} &&&&&&&&&&&&&\\
Central &&&&&& $1,256,235$ &&$921,806.1$ &&$115,000$ && $7,200,000$ \\
North &&&&&&$807,214$ &&$517,334$&&$134,306$&&$4,500,000$\\
South&&&&&&  $920,577$ &&$528,865.8$&&$243,000$ &&$4,550,000$ \\
East&&&&&&   $646,616$  &&$373,423.8$ &&$112,897$ &&$3,350,000$ \\
Gunbarrel&&&&&& $593,812$&& $307,783.2$&&$194,585$&&$1,995,051$ \\
Rural&&&&&&1,$173,444$ &&$929,252.6$&& $425,000$&&$5,779,000$\\[1ex]

\hline \hline
\end{tabular}
\end{adjustbox}
\label{table:nonlin}
\caption*{\textbf{Note}:  The house prices have been recorded in the US dollars ($\$$). }
\end{table}

As mentioned earlier, the housing market in Boulder is classified into condominiums, town-homes, and single-family houses. To account for differences in property type and location, categorical variables are added to the models using one-hot encoding. The data is then split into training, validation, and test data sets using random sampling.\\

\section{Methodology}
\begin{figure}[h!]
\centering
\includegraphics[width=\textwidth]{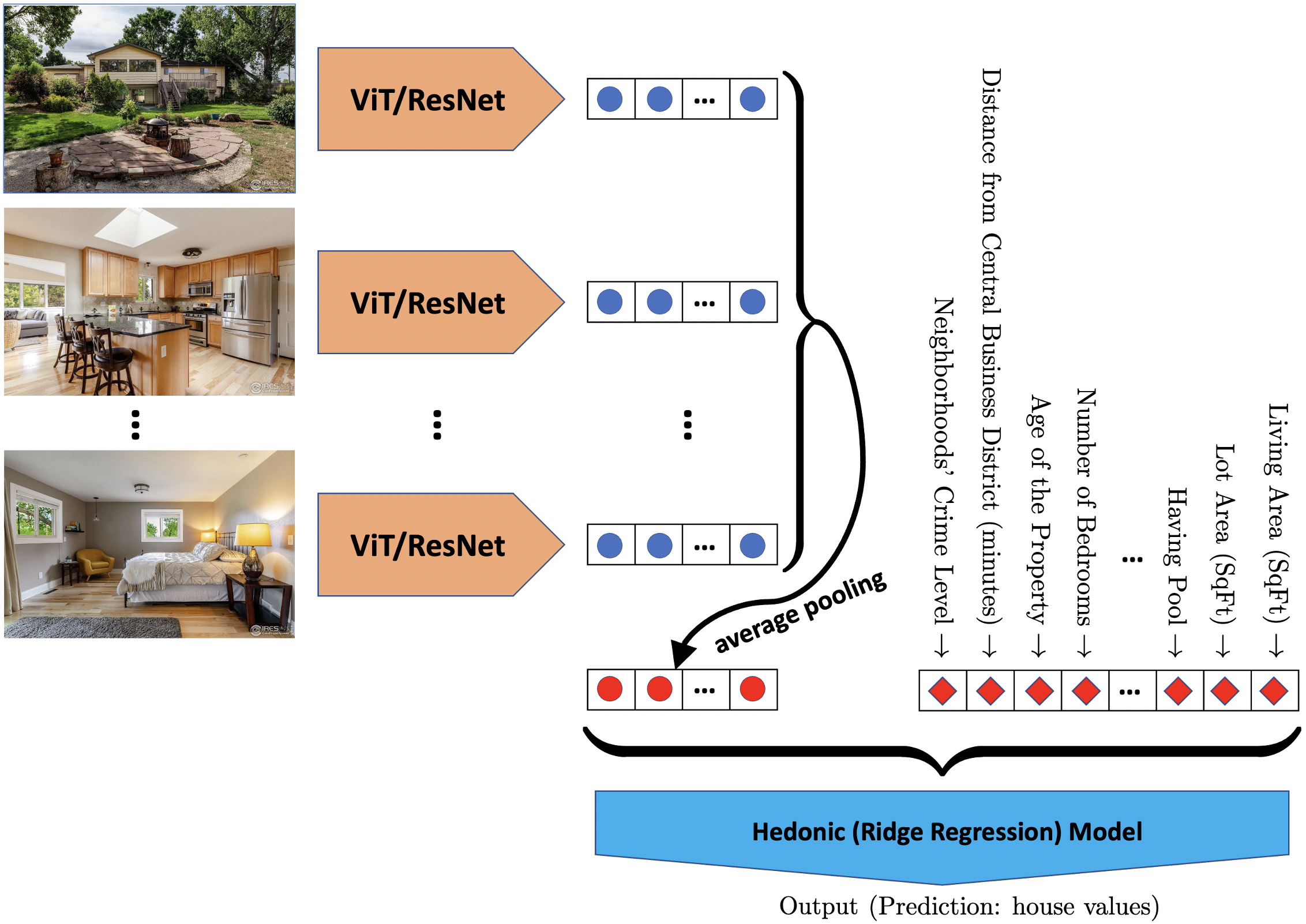}
\caption{In our data set, for each property there exist multiple images. Namely, interior images (e.g., living room, dining room, bedroom, bathroom), exterior images (e.g., house architectural style, the texture of the building material, the style of windows and doors) and street views. We take those images and extract their corresponding feature vectors by feeding them through a pre-trained Vision Transformer or a CNN (e.g., ResNet). We then aggregate the extracted feature vectors using an average pooling mechanism. We will then train a hedonic model (i.e., Ridge regression) using the pooled extracted image features and the other quantitative features (e.g., structural factors, socio-economic status of the neighborhood, environmental amenities, and location) to predict house values.}\label{main_figure_final}
\end{figure}

Our data set includes a wide variety of images for each property, including detailed interior shots of rooms like the living room, dining room, bedrooms, and bathrooms, as well as exterior images showcasing the architectural style, texture of the building materials, and the design of windows and doors. Additionally, we also have street view images which give us a sense of the surrounding neighborhood and the overall aesthetic of the area. To make the most of this wealth of information, we take these images and extract their corresponding feature vectors by feeding them through a pre-trained Vision Transformer or a CNN (e.g., ResNet). Once the feature vectors have been extracted, we then aggregate them using an average pooling mechanism. This process allows us to combine the information from all of the images and create a single, representative feature vector for each property. This is an important step because it allows us to effectively capture the most important information from all of the images in a concise and manageable format. We will then train a hedonic model (i.e., Ridge regression) using the pooled extracted image features and the other quantitative features such as structural factors, socio-economic status of the neighborhood, environmental amenities, and location. This combination of image features and quantitative data allows us to have a more complete understanding of each property, and enables us to make more accurate predictions about house values. Overall, this process of extracting, aggregating, and training on image features is a crucial step in our efforts to predict house values and gain valuable insights into the real estate market. This process is depicted in Fig.~\ref{main_figure_final}.\\

\begin{figure}[h!]
\centering
\includegraphics[width=\textwidth]{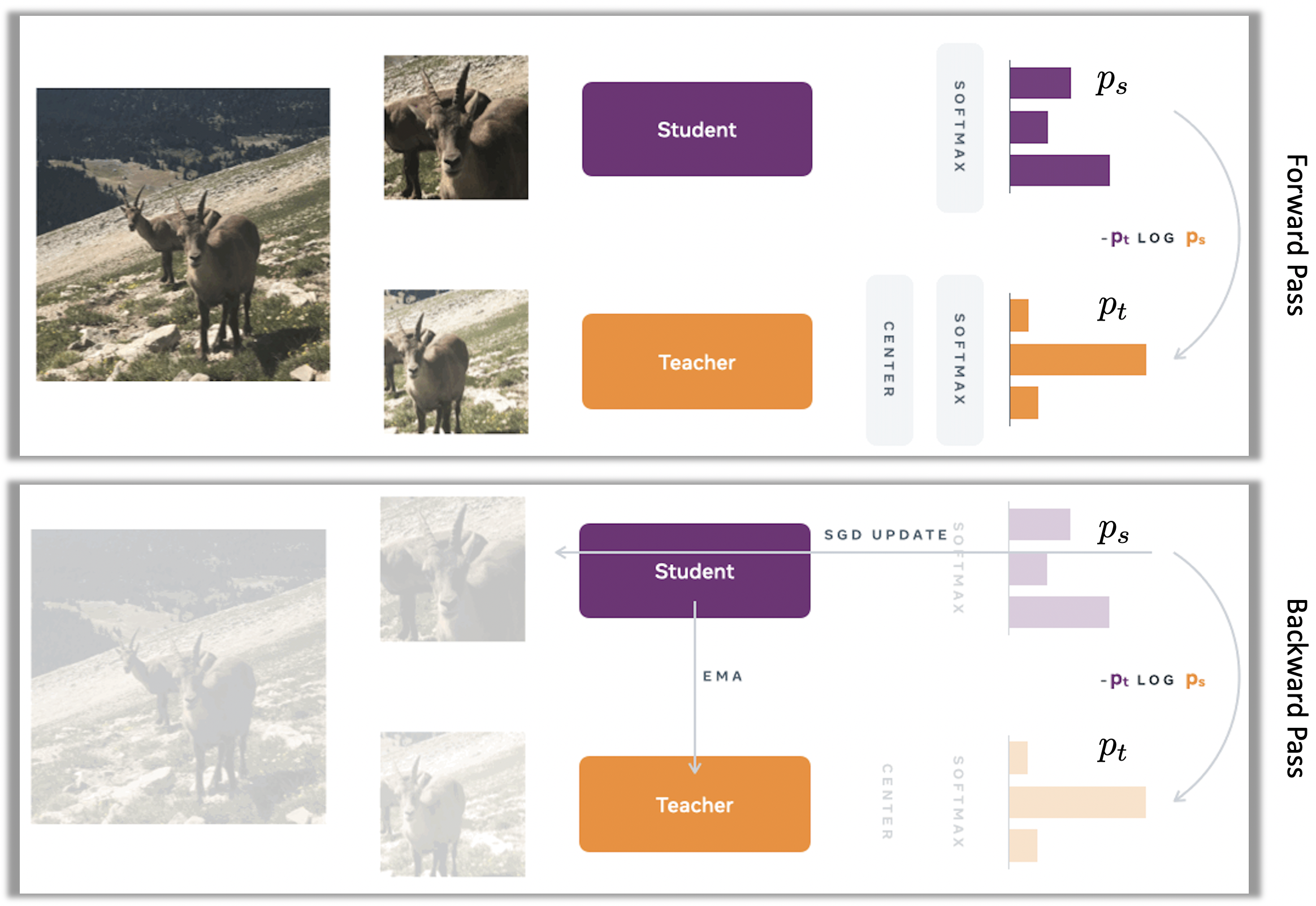}
\caption{DINO is a self-supervised approach that uses two networks, a student and a teacher, to extract features from input images. Both networks have the same architecture but different parameters. The student network is trained using stochastic gradient descent (SGD) to mimic the teacher network's output, which is measured by a cross-entropy loss. The teacher's parameters are updated with an exponential moving average (EMA) of the student's parameters.}\label{dino}
\end{figure}

To extract image features, we make use of the latest advancements in computer vision and machine learning by leveraging Vision Transformer (ViT) [\cite{dosovitskiy2020image}] and ResNet [\cite{he2016deep}] models that have been trained in a self-supervised manner on the ImageNet [\cite{russakovsky2015imagenet}] data set. This data set contains a large number of images across a wide range of categories and is widely used as a benchmark for training and evaluating computer vision models. The self-supervised learning technique used for this work is self-\underline{DI}stillation with \underline{NO} labels (DINO) [\cite{caron2021emerging}]. DINO shares the same overall structure as recent self-supervised learning approaches, such as the ones proposed in \cite{caron2020unsupervised} \cite{chen2020simple}, \cite{chen2021exploring} \cite{he2020momentum} and \cite{grill2020koray}, that have been proposed in the literature. These approaches are designed to learn visual representations from large-scale image data sets without the need for manual annotation (i.e., labeling). DINO [\cite{caron2021emerging}] also shares some similarities with knowledge distillation [\cite{hinton2015distilling}], a technique that has been widely used to improve the performance of deep neural networks by transferring knowledge from a larger and more powerful model (i.e., teacher) to a smaller and more efficient one (i.e., student). The DINO framework also utilizes two networks, a student and a teacher, to extract features from input images. Here, both networks have the same architecture but different parameters. DINO is illustrate in Fig.~\ref{dino} for simplicity with one single pair of views. However, the model actually takes multiple different random transformations of an input image and passes them to the student and teacher networks. The output of the teacher network is centered using a mean computed over the batch. Each network outputs a feature vector, which is then normalized using a temperature softmax over the feature dimension. The similarity between the student and teacher networks is measured using a cross-entropy loss. To ensure that the gradients are only propagated through the student network, a stop-gradient operator is applied on the teacher network. The teacher's parameters are updated using an exponential moving average (EMA) of the student's parameters. This approach allows for the efficient transfer of knowledge from the teacher network to the student network, ultimately leading to the improvement of the performance of the student network. This provides a clear and detailed overview of the steps involved in the framework and how they are interconnected, making it easier to understand the workings of the method and how it can be applied to different tasks. Overall, DINO is an innovative framework that combines the best of both worlds: self-supervised learning and knowledge distillation. It allows us to learn powerful visual representations from large-scale data sets. By using these pre-trained models, we can take advantage of the knowledge they have already learned from the ImageNet data and apply it to our specific task, which is image feature extraction for predicting house values.\\

\begin{figure}[h!]
\centering
\includegraphics[width=\textwidth]{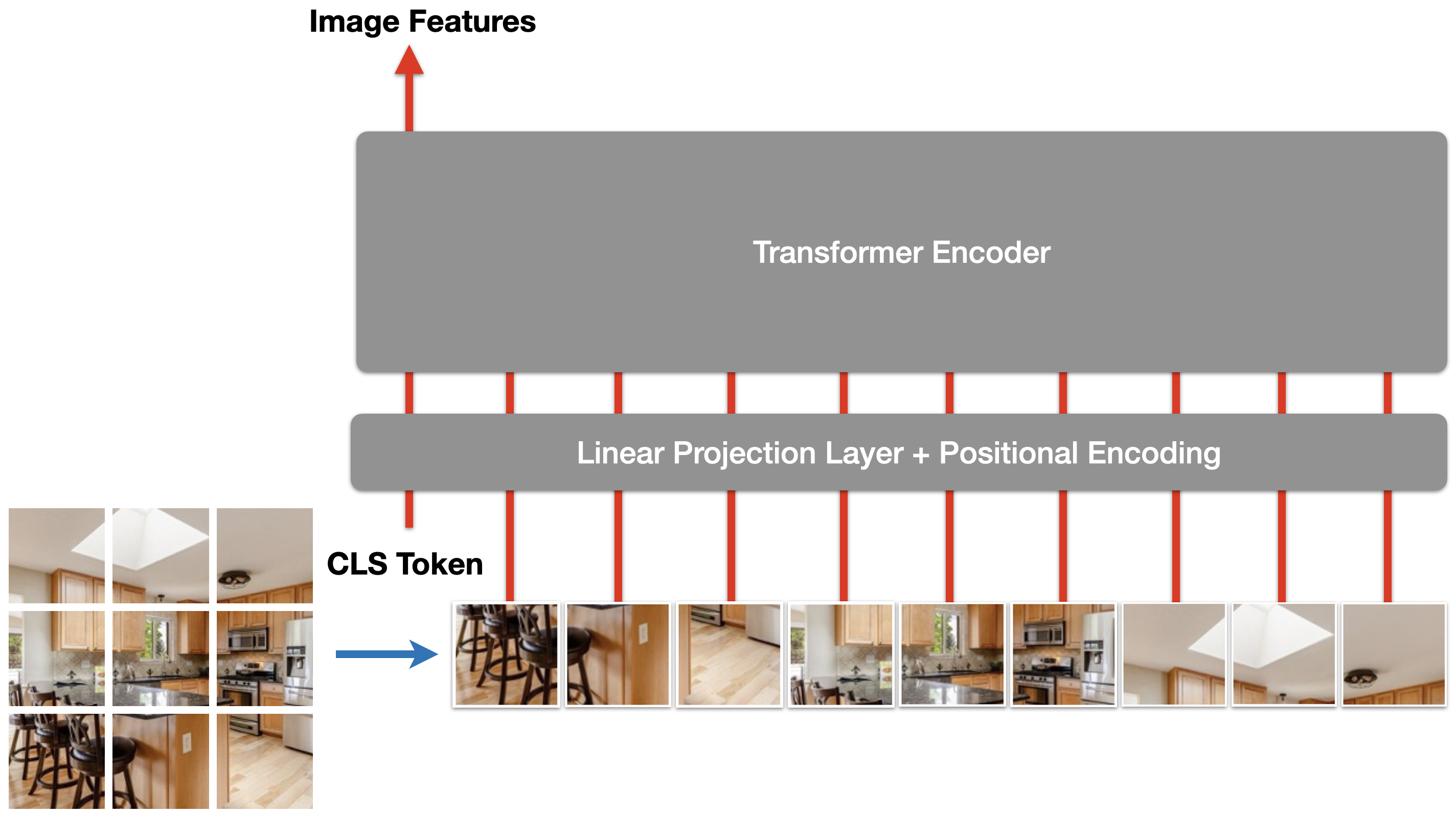}
\caption{The Vision Transformer (ViT) architecture takes an image, splits it into fixed-size patches, embeds each patch linearly, adds position embeddings, and feeds the resulting sequence of vectors to a Transformer encoder. To perform classification, an extra learnable classification (CLS) token is added to the sequence.}\label{vit}
\end{figure}

\begin{figure}[h!]
\centering
\includegraphics[width=\textwidth]{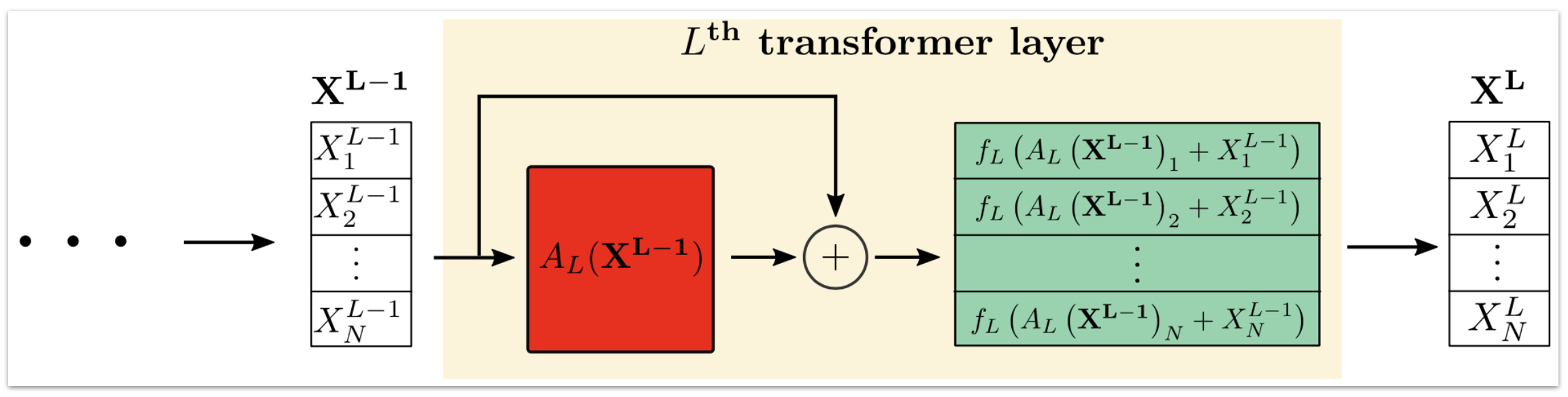}
\caption{Transformer encoders are neural network architectures that use self-attention mechanisms to weigh the importance of different parts of the input sequence when encoding it. The self-attention mechanism computes attention weights for each element in the input sequence using dot-product operation between the input elements and learnable parameters called ``keys'', ``queries'' and ``values''. These attention weights are then used to weight the input elements before they are combined to form the final encoded representation. The transformer encoder architecture also includes a multi-layer perceptron (MLP) and a residual connection, which allows the model to better capture the dependencies between the input elements. This architecture has been widely used in Natural Language and Computer Vision tasks and achieve state-of-the-art performance on many of them.}\label{transformer}
\end{figure}

The Vision Transformer (ViT) model is a new architecture that has been shown to be highly effective in self-supervised learning. On the other hand, the ResNet model is a classic architecture that has been widely used in various computer vision tasks. Both models can be trained in a self-supervised manner using the DINO framework. Given the established reputation and proven effectiveness of the ResNet architecture, it is widely understood in the field. In light of this, we will not delve into its details within this document but instead, we will focus on providing a comprehensive and in-depth explanation of the Vision Transformer, which is a cutting-edge technique in the following. The ViT architecture is based on the Transformer architecture [\cite{vaswani2017attention}], originally developed for natural language processing, but it has been adapted for computer vision. As illustrated in Fig.~\ref{vit}, the key idea behind ViT is to split an image into fixed-size patches, linearly embed each of them, add position embeddings, and feed the resulting sequence of vectors to a standard Transformer encoder [\cite{vaswani2017attention}], this allows the model to maintain a consistent representation of the input image. To extract features using the ViT architecture, we use the standard approach of adding an extra classification (CLS) token to the sequence, this token is then used as input to a linear layer, which produces the final output of the model. The key insight of the Transformer architecture as depicted in Fig.~\ref{transformer} is that it allows the model to process input sequences in parallel, which greatly improves the model's ability to handle long-distance dependencies.\\

Transformer Encoders (see Fig.~\ref{transformer}) are neural network architectures that were introduced in the 2017 paper ``Attention is All You Need'' by Google researchers [\cite{vaswani2017attention}]. The key innovation in this architecture is the use of self-attention mechanisms, which allow the model to weigh the importance of different parts of the input sequence when encoding it. The self-attention mechanism works by computing a set of attention weights for each element in the input sequence, which indicate how much each element should be taken into account when encoding the sequence. These attention weights are computed using a dot-product operation between the input elements and a set of learnable parameters called ``keys'', ``queries'' and ``values''. The dot-product scores are then passed through a softmax function to obtain the attention weights, which are used to weight the input elements before they are combined to form the final encoded representation. The transformer encoder architecture also includes a multi-layer perceptron (MLP) and a residual connection, which allows the model to better capture the dependencies between the input elements. This architecture has been used in a variety of natural language processing (NLP) tasks, such as machine translation, text summarization, and language modeling, and has achieved state-of-the-art performance on many of them. The ViT architecture also relies on Transformer Encoders and their attention mechanism.

\section{Results}
This section compares the performance of various computer vision architectures, trained in a self-supervised manner using the DINO framework, as image feature extractors for transfer learning (see Table \ref{main_table}). The performance of each model is measured using Root Mean Square Error (RMSE), which is a widely used metric for evaluating the performance of predictive models. This comparison is made against a baseline hedonic model (i.e., Ridge regression) that only uses quantitative features such as number of bedrooms, bathrooms, square footage, lot size, property age, crime rates, and proximity to amenities. In contrast, the other models in Table \ref{main_table} include both image features extracted from various computer vision models and the aforementioned quantitative features. This combination of features provides a more comprehensive view of the data, which can lead to improved predictions. The ``Alpha'' column in Table \ref{main_table} displays the constant that multiplies the L2 term in Ridge regression, which is a classical linear model. This constant controls the strength of regularization, which helps prevent overfitting, a common problem in machine learning where a model becomes too closely fitted to the training data and fails to generalize well to new, unseen data. The ``Improvement over Baseline'' column in Table \ref{main_table} shows the percentage improvement in RMSE over the baseline for each of the other models. The results in this column demonstrate that all models incorporating image features perform better than the baseline model, as indicated by their lower RMSE values on the test data. Of all the architectures in the table, the one with the best performance is ViT-B/8, with an improvement of 10.63\% over the baseline. This highlights the potential of computer vision models in transfer learning, as they can be used to extract meaningful image features that can be combined with other features to improve the accuracy of predictive models.\\

\begin{table}[h!]
\centering
\begin{adjustbox}{max width=\textwidth}

{\begin{tabular}{|c|c|c|c|c|c|}
\hline
Architecture &  Train-RMSE  &  Val-RMSE & Test-RMSE &Alpha& Improvement over Baseline\\
\hline
Baseline &\$117.28& \$130.34& \$117.09 &40& 0.00\%\\
\hline
ResNet-50&\$57.57& \$130.93& \$108.85 &360&7.00\%\\
\hline
ViT-B/16&   \$77.07& \$126.48& \$106.62  &  350  &8.94\%\\
\hline
ViT-S/16&\$86.60& \$124.30& \$106.15 &290&9.34\%\\
\hline
ViT-S/8&  \$75.63& \$119.79& \$105.74    &100& 9.69\%\\

\hline
ViT-B/8& \$73.88 &\$126.23 & \$104.64 & 320&10.63\%\\

\hline

\end{tabular}}
\end{adjustbox}
\caption{This table compares the performance of various computer vision architectures as image feature extractors for transfer learning using Root Mean Square Error (RMSE) as the evaluation metric. The baseline hedonic model (Ridge regression) serves as a comparison, using only quantitative features such as number of bedrooms, bathrooms, square footage, lot size, property age, crime rates, and proximity to amenities. In contrast, the other models incorporate both the extracted image features and the quantitative features. The results show that the baseline architecture has the highest RMSE on the test data, while the other models perform better, with lower RMSE values. The ``Alpha'' column displays the constant multiplying the L2 term in Ridge regression, which controls regularization strength, while the ``Improvement over Baseline'' column shows the improvement in RMSE in percentage over the baseline architecture. Out of all the architectures, ViT-B/8 achieves the best performance with the lowest RMSE on the test data set and an improvement of 10.63\% over the baseline.}\label{main_table}
\end{table}

Table \ref{vit_table} provides information about the configurations of different computer vision architectures used as image feature extractors in this work. The columns in the table are labeled ``Blocks'', ``Dim'', ``Heads'', ``\# Tokens'', ``\# Params (M)'', and ``Im/''. The ``Blocks" column refers to the number of Transformer blocks in the network. The ``Dim'' column refers to the channel dimension of the network. The ``Heads'' column represents the number of heads in the multi-head attention mechanism. The ``\# Tokens'' column indicates the length of the token sequence when the network is fed with inputs of a resolution of $224 \times 224$ center-cropped from property images. The ``\# Params'' column specifies the total number of parameters in the network, excluding the projection head \cite{caron2021emerging}. Finally, the ``Im/s'' column lists the inference time of the network on a NVIDIA V100 GPU, with 128 samples processed in each forward pass. The table is intended to provide a clear and concise overview of the network configurations, allowing readers to easily compare and understand the differences between the different models. The ViT architecture takes as input a grid of non-overlapping contiguous image patches of resolution $N \times N$. In this paper, $N = 16$ (``/16'') or $N = 8$ (``/8''). In Tables \ref{main_table} and \ref{vit_table}, ``-S'' refers to ViT small and ``-B'' indicates the ViT base architecture. The findings presented in this paper (see Table \ref{main_table}) align with the previously published research (see e.g., \cite{caron2021emerging}), which shows that models with a larger size using images divided into smaller patches (e.g., ViT-B/8) tend to have better performance. Moreover, all ViT models outperform ResNet despite being trained using the same self-supervised technique, namely DINO. One reason why ViT may perform better than ResNet is its use of the self-attention mechanism. Unlike traditional convolutional neural networks (CNNs) such as ResNet, ViT employs self-attention mechanisms to directly model relationships between all elements in the input sequence, rather than just neighboring elements. This allows ViT to capture more complex and global dependencies in the input data, resulting in improved performance. However, it should be noted that all ViT models are slower feature extractors than ResNet as illustrated in Table \ref{vit_table}.\\

\begin{table}[h!]
\centering
\begin{tabular}{|c|c|c|c|c|c|c|}
\hline
Model & Blocks & Dim & Heads & \# Tokens & \# Params (M) & Im/s \\
\hline
ResNet-50 & - & 2048 & - & - & 23 & 1237 \\
\hline
ViT-S/16 & 12 & 384 & 6 & 197 & 21 & 1007 \\
\hline
ViT-S/8 & 12 & 384 & 6 & 785 & 21 & 180 \\
\hline
ViT-B/16 & 12 & 768 & 12 & 197 & 85 & 312 \\
\hline
ViT-B/8 & 12 & 768 & 12 & 785 & 85 & 63 \\
\hline
\end{tabular}
\caption{This table outlines the configurations of different networks. It lists the number of Transformer blocks as ``Blocks'', the channel dimension as ``Dim'', and the number of heads in multi-head attention as "Heads". The length of the token sequence for inputs with a resolution of 224x224 is listed as ``\# Tokens''. The total number of parameters (excluding the projection head) is listed as ``\# Params (M)'' in million. The ``Im/s'' column shows the time taken for one forward pass of 128 samples on an NVIDIA V100 GPU [\cite{caron2021emerging}].}\label{vit_table}
\end{table}

\begin{figure}[h!]
\centering
\includegraphics[width=\textwidth]{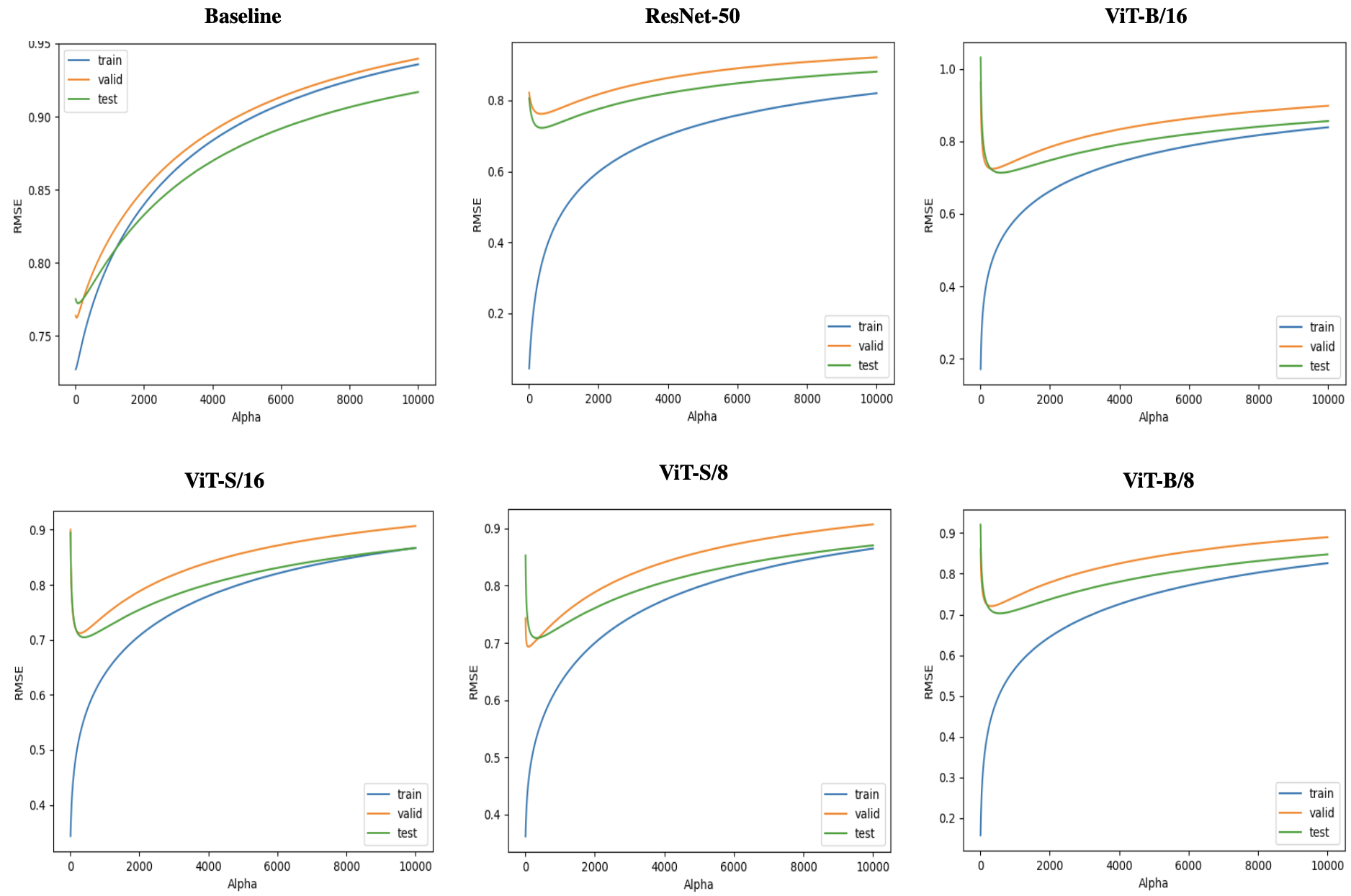}
\caption{The best hedonic model is chosen based on the hyper-parameter ``Alpha'' (i.e., the hyper-parameter controlling the regularization strength) that results in the lowest RMSE on the validation data.}\label{overfitting}
\end{figure}

The property images are transformed using computer vision techniques and used as additional inputs with quantitative features like number of rooms, square footage, age, crime rates, etc. These combined features are fed into a hedonic model (Ridge Regressor) to predict the property value. Incorporating image features increase the total number of variables, and as a consequence the number of parameters in the hedonic model, and make it prone to overfitting, which is why we use validation data to determine the strength of regularization through the constant multiplying the L2 term in Ridge regression. This helps prevent overfitting. The best model is chosen based on the hyper-parameter ``Alpha'' (i.e., the hyper-parameter controlling the regularization strength) that results in the lowest RMSE on the validation data, as shown in Fig.~\ref{overfitting}.\\

The RMSE numbers in Table \ref{main_table} are reported in dollars because the purpose of the model is to estimate the value of real estate properties. The RMSE is a measure of the difference between the predicted value and the actual value of a property. When the RMSE is reported in dollars, it provides a clear and intuitive understanding of the magnitude of the error in the prediction. For example, an RMSE of \$100 means that on average, the model's predictions are off by \$100 per square footage. This means that the model's prediction error for a property with 2,000 square feet of living space would be \$100 $\times$ 2,000 $=$ \$200,000. A 2,000 square feet residential property in Boulder, CO could worth above \$2,000,000. A reduction in the RMSE of \$1 per square footage, by a more accurate model, would mean that the average prediction error for a property with 2,000 square feet of living space would decrease from \$100 $\times$ 2,000 = \$200,000 to \$99 $\times$ 2,000 = \$198,000. This leads to a \$2,000 difference in evaluation. This can have important implications for the real estate industry, as it can result in more accurate pricing and better informed decisions for buyers, sellers, and lenders.\\

This work proposes a new AI-based method for property valuation in real estate. The use of self-supervised vision transformers, machine learning, computer vision, and hedonic pricing models trained on real estate data is expected to improve the accuracy of property value estimation, outperforming traditional appraisal methods. The method has potential for real-world applications and its significance lies in the importance of accurate property valuation for the functioning of the real estate market. Improved property valuation methods can result in more efficient and fair transactions and better investment decisions. The use of AI in property valuation can have a positive impact on the real estate market and the economy as a whole.

\section{Concluding Remarks and Future Works}
In conclusion, this paper proposed a new method for property valuation utilizing self-supervised vision transformers, a recent breakthrough in computer vision and deep learning. The proposed algorithm uses a combination of machine learning, computer vision and hedonic pricing models trained on real estate data to estimate the value of a given property. We collected and pre-processed a data set of real estate properties in the city of Boulder, Colorado and used it to train and test our algorithm. Our data set consisted of qualitative images as well as quantitative features such as the number of bedrooms, bathrooms, square footage, lot square footage, property age, crime rates, and proximity to amenities. We evaluated the performance of our model using metrics such as Root Mean Squared Error (RMSE). Our findings indicate that these techniques are able to accurately predict the value of properties, with a low RMSE. The proposed algorithm outperforms traditional appraisal methods that do not leverage property images and has the potential to be used in real-world applications. The use of AI in the real estate industry is growing in recent years, and our research highlights the potential for self-supervised vision transformers to revolutionize the property valuation process. With continued development and refinement, this algorithm could become a valuable tool for real estate professionals, making the process of property valuation more efficient and accurate. Additionally, this research is a step towards creating more fair and accurate models for property valuation that are not susceptible to human bias. We believe that our proposed algorithm has the potential to make a significant impact on the real estate industry and we look forward to seeing it being used in real-world applications.\\

In future work, making use of data sets from different regions and cities for property valuation will be crucial in enhancing the generalizability and accuracy of the proposed algorithm. Fine-tuning the model to these data sets could further improve its performance. Implementing the algorithm in real-world scenarios and gathering feedback from real estate professionals will offer valuable insights into its practicality and efficacy. Furthermore, incorporating other computer vision techniques such as object detection and semantic segmentation is also a potential direction. Additionally, leveraging textual data such as property descriptions can also be explored. The proposed algorithm has the potential to revolutionize the property valuation process, but further research is necessary to fully tap into its potential.

\renewcommand\bibname{References}

\bibliographystyle{apalike}

\bibliography{mybib.bib}
\newpage

\clearpage
\pagenumbering{arabic}% resets `page` counter to 1
\renewcommand*{\thepage}{A\arabic{page}}
\renewcommand\appendixtocname{Appendices}

\clearpage
\appendix
\renewcommand\thefigure{\thesection.\arabic{figure}} 
\renewcommand\thetable{\thesection.\arabic{table}}

\label{app:Online}

\end{document}